\documentclass[11pt]{article}
\usepackage[preprint]{acl}
\usepackage{times}
\usepackage{latexsym}
\usepackage[T1]{fontenc}
\usepackage[utf8]{inputenc}
\usepackage{microtype}
\usepackage{graphicx}
\usepackage{booktabs}
\usepackage{amsmath,amssymb}
\usepackage{tabularx}
\usepackage{xcolor}
\newcommand{\revisioncolor}{black}
\newcommand{\revision}[1]{\textcolor{\revisioncolor}{#1}}
\newcommand{\method}{FinEvo}
\newcommand{\pct}{\%}
\title{FinEvo: Recursive Self-Improvement for\\Long-Horizon Financial Modeling}
\author{
Chengjun Pan$^1$ \quad Shichun Liu$^2$ \quad Zekai Zhou$^1$ \quad Shihan Dou$^2$\\
\textbf{Jiahang Lin$^2$ \quad Chengsong Huang$^3$ \quad Dingwei Zhu$^2$ \quad Tao Gui$^2$}\thanks{Corresponding author: \texttt{tgui@fudan.edu.cn}.}\\
\textnormal{$^1$Peking University \qquad $^2$Fudan University}\\
\textnormal{$^3$Washington University}
}
\hypersetup{
  pdftitle={FinEvo: Recursive Self-Improvement for Long-Horizon Financial Modeling},
  pdfauthor={Chengjun Pan, Shichun Liu, Zekai Zhou, Shihan Dou, Jiahang Lin, Chengsong Huang, Dingwei Zhu, Tao Gui}
}
\begin{document}
\maketitle

\begin{abstract}
Advances in large language models (LLMs) have expanded their role in high-value productivity tasks, including financial analysis and modeling, while enabling models to play an increasingly active role in their own improvement.
This motivates two complementary lines of inquiry: can financial modeling capabilities provide a foundation for model self-evolution, and can self-evolution, in turn, advance financial capabilities?
We introduce \textbf{\method{}}, a self-evolution framework that integrates autonomous experience construction with financial capability improvement.
FinEvo prepares financial models, iteratively evolves challenges through self-play, and learns from the resulting interactions using executable rewards, relying on the same model throughout.
\revision{An end-to-end execution with Qwen3.6-27B shows that solving feedback supports both increasing reasoning demands and identifying specification issues that executable checks can miss.}
\revision{Learning from these experiences raises whole-task success from \textbf{22.73\pct{} to 29.09\pct{}} on held-out within-source tasks and from \textbf{13.13\pct{} to 19.19\pct{}} on SpreadsheetBench 2 Financial.}
\revision{Comparable gains across task sources and evaluation settings, together with improvement on a task category absent from training, suggest that FinEvo strengthens underlying financial modeling capabilities shared across tasks. These findings support self-evolution as a path toward broader financial competence, advancing the empirical exploration of recursive self-improvement (RSI) in professional domains.}
Code is available at \url{https://github.com/pcj-sudo/FinEvo}.
\end{abstract}

\begin{figure*}[t]
\centering
\includegraphics[width=\textwidth]{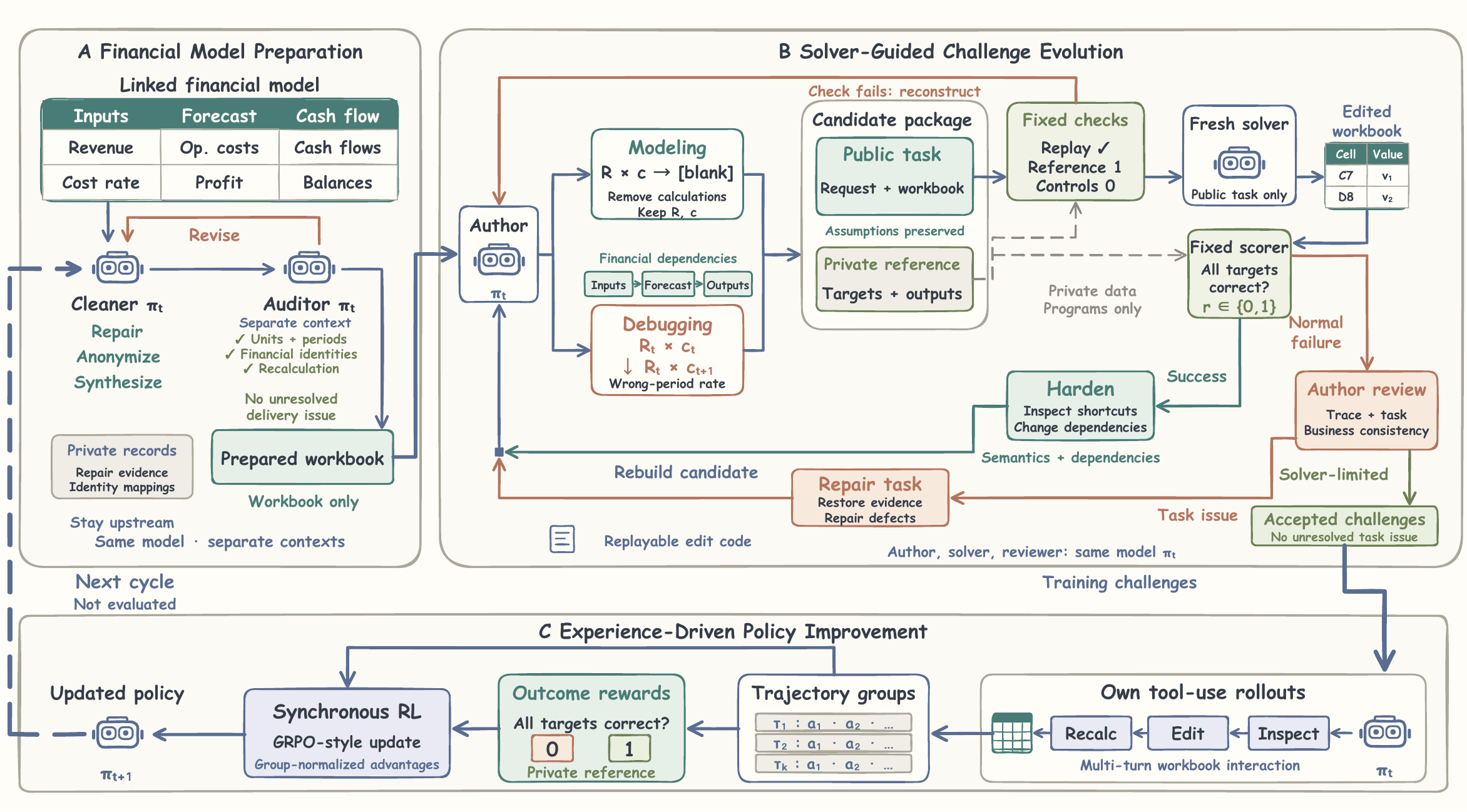}
\caption{Overview of \method{}. All model roles use the same underlying model. (A) Financial Model Preparation iterates cleaning and independent auditing, releasing only a prepared workbook while retaining private cleaning records upstream. (B) Solver-Guided Challenge Evolution checks candidates, gives fresh solvers only the public task, and reserves private references for fixed programs. Successful solves guide hardening; reviewed failures lead to task repair or qualified acceptance. (C) Experience-Driven Policy Improvement learns from the policy's own tool-use rollouts using binary whole-task rewards. Transformations and cell values are schematic ($R$: revenue; $c$: cost rate; $t$: period). The dashed return illustrates recursive reuse; experiments evaluate one generation--learning cycle on fixed prepared sources.}
\label{fig:workflow}
\end{figure*}

\section{Introduction}

The capabilities and applications of large language models (LLMs) have expanded from single-turn question answering, multistep reasoning and code generation to high-value professional tasks requiring sustained tool use and delivery of complete artifacts \citep{deepseekmath,openhands,coworkos}.
These capabilities are also being used to advance AI itself: AI-for-AI (AI4AI) work explores how models can accelerate machine learning research and development \citep{mlmaster}, and some studies have even begun to investigate the possibility of recursive self-improvement (RSI), using existing capabilities to support further improvement \citep{frontisma1}.
Such studies report gains in reasoning, coding and interactive decision-making, providing evidence that models can contribute to improving their own capabilities \citep{absolutezero,mae,stretch}.

Financial analysis and modeling are among the professional capabilities receiving particular attention in frontier model development \citep{anthropicfinance,kimiK31Pro}; they support forecasting, valuation and investment decisions \citep{cfainstitute} and offer substantial productivity value \citep{frontierfinance,spreadsheetbench2}.
These activities span diverse business settings and require models to combine domain knowledge with reasoning about complex relationships between business assumptions and financial calculations \citep{cfainstitute,icaewmodellingcode}.
Training for these capabilities raises the question of how to develop learning experiences that cover diverse settings while matching the model's level of ability.
Research on automated curricula and self-play has shown that organizing training challenges around learner performance can improve learning \citep{graves2017curriculum,sukhbaatar2018selfplay,envharness}; recent work exploring RSI has further demonstrated the potential for models to improve by autonomously generating and learning from experience \citep{absolutezero,mae,stretch}.
Together, these findings motivate exploring a path for financial capability development in which models' own capabilities and experience help shape their subsequent learning.

We therefore ask two complementary research questions:
\textit{RQ1: In financial analysis and modeling, can a model use its existing capabilities and feedback from solving tasks to autonomously construct and revise experiences that support its own learning?}
\textit{RQ2: Can training on the learning experiences that a model autonomously constructs and revises further improve its financial analysis and modeling capabilities?}

To investigate these questions, we introduce \textbf{\method{}}, a self-evolution framework for financial analysis and modeling organized into three stages (Figure~\ref{fig:workflow}).
\emph{Financial Model Preparation} cleans and audits financial models to establish the basis for task construction.
\emph{Solver-Guided Challenge Evolution} iteratively evolves financial challenges through self-play between task-author and solver roles of the same model.
\emph{Experience-Driven Policy Improvement} trains the target policy on its own interactions with the generated challenges using executable outcome rewards.
Together, these stages rely entirely on the same model's own capabilities to prepare financial models, evolve challenges, and learn from its own experience.

Our experiments yield two complementary findings.
\emph{Solver feedback supports both increasing reasoning demands and revising the learning tasks themselves.}
Construction histories show that analyzing successful solves helps the model develop challenges with greater business-semantic and dependency-reasoning demands, while reviewing failures reveals missing public assumptions and specification issues even after executable checks pass (RQ1; Section~\ref{sec:generation}).
\revision{\emph{Learning from self-generated challenges generalizes across task sources and types.}}
\revision{RL yields comparable whole-task gains on within-source tasks (+6.36 percentage points) and SpreadsheetBench 2 Financial (+6.06), despite changes in task sources and evaluation settings. Template filling improves most (+13.40 points), although RL uses only modeling and debugging challenges (RQ2; Section~\ref{sec:learning}). Together, these results support improvements in underlying financial modeling capabilities shared across tasks, demonstrating the training value of self-generated challenges beyond their source workbooks and task types.}

Our contributions are threefold.

\paragraph{(1) A self-evolution framework for financial analysis and modeling.}

We introduce \method{}, which couples financial challenge evolution with policy improvement, closing the loop between constructing learning experiences and advancing financial capabilities.

\paragraph{(2) Financial challenges with controlled difficulty and validity.}

We construct over 100 financial modeling and debugging challenges, combining iterative hardening with executable checks and solver-informed review to strengthen both reasoning demands and task validity.

\paragraph{(3) An empirical step toward RSI in financial tasks.}

\revision{Our complete construction--learning execution with Qwen3.6-27B yields financial capability gains that generalize across task sources and types, contributing to the empirical exploration of RSI.}

\section{Preliminaries}
\label{sec:preliminaries}
\label{sec:sources}

\paragraph{Financial models.}
A financial model is an executable representation of business relationships in a spreadsheet workbook. Input assumptions and historical data drive formulas for operating forecasts, linked financial statements, valuation, financing or contract schedules. For example, revenue and a cost-rate assumption determine operating costs, which propagate into profit and cash flow. Worksheets encode these relationships through formulas, labels, units, dates and notes. Complete models provide both calculation structure and reference results. Known assumptions are distinct from quantities to be inferred: a formula may carry an independent assumption, while a hardcoded value may be a derived result.

\paragraph{Modeling and debugging tasks.}
We consider two task types. \emph{Modeling} requires completing missing financial calculations from the available business assumptions and workbook structure. \emph{Debugging} requires locating and correcting errors in existing calculations. In both cases, the agent receives a task description and an input workbook, then submits an edited workbook whose outputs are evaluated against a private reference. Task completion requires interpreting business meanings, units and periods, tracing calculation dependencies across worksheets, and inspecting, editing and recalculating the workbook through multistep tool interaction.

\begin{figure*}[t]
\centering
\includegraphics[width=\textwidth]{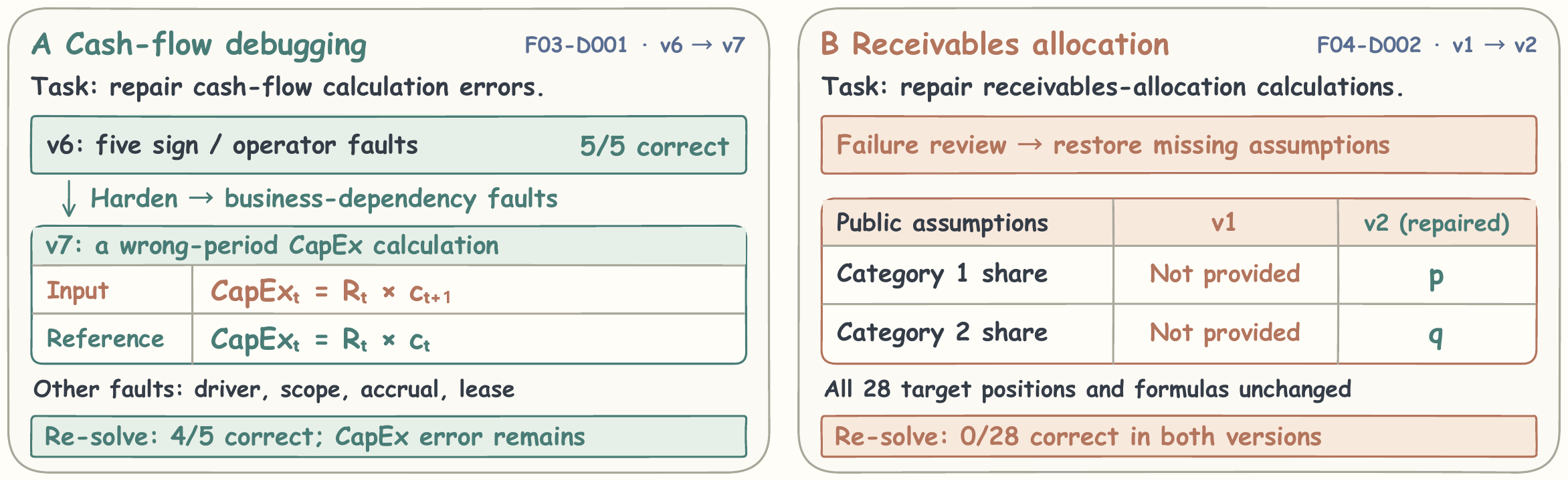}
\caption{Concrete examples of generated tasks and their evolution. Task descriptions are paraphrased; workbook excerpts are schematic. (A) Cash-flow debugging is hardened from five sign/operator faults to five business-dependency faults; the wrong-period CapEx rate remains after re-solving. (B) Review restores missing public allocation labels and ratios while all 28 targets and formulas remain unchanged; both solves score zero. $R$: revenue; $c$: CapEx rate; $t$: period; $p,q$: restored ratios. Appendix~\ref{app:cases} gives the retained evidence.}
\label{fig:cases}
\end{figure*}

Figure~\ref{fig:cases} illustrates concrete debugging tasks and how solver feedback changes their difficulty or public evidence.

\section{FinEvo}
\label{sec:method}

\method{} integrates autonomous experience construction with policy improvement through three stages (Figure~\ref{fig:workflow}). \emph{Financial Model Preparation} establishes reference workbooks for task construction. \emph{Solver-Guided Challenge Evolution} transforms these workbooks into modeling and debugging challenges, using solving feedback to guide iterative hardening and repair. The accepted challenges then define the training distribution for \emph{Experience-Driven Policy Improvement}, where the policy learns from its own interactions using executable outcome rewards. The same model performs preparation, challenge evolution and learning.

\subsection{Financial Model Preparation}
\label{sec:cleaning}

Figure~\ref{fig:workflow}A prepares a financial model for downstream task construction: a cleaner produces a revised workbook, an auditor checks it in a separate context, and unresolved issues return for revision. Both roles use the same target model and maintain their own persistent contexts. The cleaner first identifies business definitions, units, periods, formula dependencies and accounting relationships in the source materials. It repairs calculation errors, conflicting definitions and missing assumptions while preserving supported business choices.

The cleaner\textquotesingle{}s three operations in the figure are repair, anonymization and value synthesis. Anonymization replaces real company, project, institution, person and contract names, including residual identifiers in comments, document properties, hidden sheets and external links. Value synthesis creates fictional business values from base inputs and business drivers, then recalculates dependent forecasts and outputs; independent cell perturbations or uniform scaling do not suffice. Replayable code records these operations, with repair decisions and their evidence distinguished from anonymization and synthetic-data transformations. Necessary business definitions remain in ordinary assumption sheets, headers, notes or comments.

The auditor then independently inspects the resulting workbook before consulting source materials and internal records to verify repairs and transformations. Recalculation and independent computations check financial relationships, units, timing and dependencies. As the return arrow in Figure~\ref{fig:workflow}A indicates, unresolved issues trigger another cleaner--auditor round; budget exhaustion stops preparation. The prepared workbook is released only when no unresolved delivery issue remains. Only this workbook enters challenge evolution in panel B; private cleaning analyses, repair guides and identity mappings stay upstream. These checks provide a documented basis for construction, not an exhaustive guarantee of financial correctness.

\subsection{Solver-Guided Challenge Evolution}

\paragraph{Candidate construction.}
Figure~\ref{fig:workflow}B starts from the prepared workbook delivered by panel A. The same-model author selects business regions and dependencies and writes replayable transformations. Its two branches, ``Modeling'' and ``Debugging,'' respectively remove derived calculations for reconstruction or inject implementation errors for diagnosis and repair. The resulting ``Candidate package'' contains a public request and input workbook, a private reference and a private target set $T$. Transformations preserve the public assumptions needed for solving; a new business module requires its own assumptions, reference calculation and checks.

\paragraph{Fixed checks and solver feedback.}
\label{sec:scoring}
The candidate next follows ``Fixed checks'' $\rightarrow$ ``Fresh solver'' $\rightarrow$ ``Edited workbook'' $\rightarrow$ ``Fixed scorer'' in Figure~\ref{fig:workflow}B. Checks require replayed edits to match the target set, the reference to score one, and unchanged input and deliberately wrong outputs to score zero. A failed check follows the upper return arrow to the author for reconstruction. After passing, a fresh same-model solver receives only the public task; the private reference and target list feed the fixed checking and scoring programs, as shown by the separate gray path. Scoring compares values after the same LibreOffice recalculation and normalization, using two decimal places for numbers and normalized dates. For task $q$ and attempt $j$,
\begin{align}
 a_{qj} &= |T_q|^{-1}\sum_{c\in T_q}\mathbf{1}[v_{qj}(c)=v_q^*(c)],\\
 r_{qj} &= \mathbf{1}[a_{qj}=1].
\end{align}
Here $v_q^*$ denotes reference values. Missing or unreadable workbooks receive zero; equivalent formulas or correct values are accepted. Non-target changes are diagnostic, although they may propagate into target errors. Debugging targets are direct injected-error positions. The scorer and its numerical rule remain fixed.

\paragraph{Hardening, review and acceptance.}
The two outcome paths leave the fixed scorer. The ``Success'' path reaches ``Harden'': the author reads the successful trajectory, identifies shortcuts, and constructs a harder successor through cross-sheet dependencies, interacting faults or less reusable formula patterns. The ``Normal failure'' path reaches ``Author review,'' which examines the trajectory, output and target differences against public evidence and business relationships. Ambiguity, missing assumptions or reference defects lead to ``Repair task.'' Only failure attributed to solver error, with business consistency checked and no unresolved task issue, reaches ``Accepted challenges.'' A failed attempt alone therefore does not establish a valid challenge.

The ``Harden'' and ``Repair task'' paths merge into the left-hand return arrow to the author, rebuilding the next candidate before checks and solving repeat. The author retains its task history, while each solver starts fresh. Content changes create versions; execution or API failures instead trigger bounded retries or recovery of the same frozen version.

\subsection{Experience-Driven Policy Improvement}

Reinforcement learning (RL) updates an agent policy from rewards obtained through environment interactions, with the objective of increasing expected task reward. Group Relative Policy Optimization (GRPO) \citep{deepseekmath} samples multiple trajectories per task and uses their group-normalized rewards to estimate advantages without training a separate value model.

In the RL stage of Figure~\ref{fig:workflow}C, accepted challenges define the training distribution. For long trajectories, we use a GRPO-style update adapted from the Modal Dojo/Slime community recipe \citep{modaldojo,slime} for multi-turn agents. The current policy samples multiple tool-use trajectories per task, inspecting and editing workbooks before submission. The fixed scorer assigns binary whole-task rewards. Each trajectory's reward is divided equally across its context segments, and the resulting rewards are normalized over valid segments within each task group to obtain policy advantages; reference solutions are used by the scorer, not supplied as demonstrations. Learning therefore uses the target model\textquotesingle{}s own interactions on its generated tasks.

We adopt synchronous RL to avoid policy-lag-induced off-policy mismatch: trajectories generated by an outdated policy may not reflect the behavior of the policy being optimized. Aligning data collection with the current policy makes learning feedback correspond to its present capabilities. \revision{Held-out within-source tasks assess the resulting policy with the same scoring contract, using mean whole-task success and macro-averaged target accuracy.} Together, autonomous task construction, learning from self-generated interactions and held-out improvement assessment close the minimal RSI loop evaluated in this work. The same model\textquotesingle{}s own challenge generation and executable feedback thus become the basis for improving its capabilities.

\section{Experiments}
\label{sec:experiments}

\subsection{Experimental Setup}
\label{sec:design}

\paragraph{Source workbooks.}
We use 11 internal financial models as source workbooks.

\paragraph{Generation campaign.}
Each of 11 workbooks seeds five modeling and five debugging tasks. Qwen3.6-27B \citep{qwen36} with OpenHands SDK 1.47.0 \citep{openhands} performs construction, hardening, isolated solving and review at \texttt{xhigh} reasoning. Each call has 3,600 seconds and an initial cap of ten candidate versions.

\paragraph{RL training and evaluation.}
For RL, one of the five tasks in each source--type group is held out, yielding 88 training and 22 development tasks. Training follows the adapted Modal Dojo/Slime recipe \citep{modaldojo,slime}. Each update samples 32 distinct training queries with eight fresh trajectories each, using the current policy only and admitting no mixed-version batches. The learning rate is $4\times10^{-6}$, with no KL or entropy penalty. Each checkpoint is evaluated on all 22 development tasks with five attempts per task, temperature 0.6 and a 1,800-second limit. Missing outputs score zero.

\paragraph{\revision{External transfer evaluation.}}
\revision{We evaluate the model before and after RL on all 297 tasks in the Financial package of SpreadsheetBench 2 \citep{spreadsheetbench2}: 100 financial modeling, 97 template filling and 100 error fixing tasks. We use OpenHands and report whole-task success using the official scoring code.}

\paragraph{Independent solvability check.}
We conduct a small-scale supplementary solvability check on eight randomly selected tasks generated by \method{}. GPT-6/Codex solves the same tasks in a separate harness. This comparison provides no solutions or supervision for task acceptance or RL.

\subsection{Self-Generated Training Challenges (RQ1)}
\label{sec:generation}

We analyze whether repeated construction, checking and independent solving produce valid tasks at the target difficulty. Success triggers hardening; failures prompt review of task validity using trajectories and outputs. Identified task issues lead to revision and another solve, while failures attributed to solver limitations are accepted after review. We analyze round-wise outcomes, hardening and repair paths to characterize the resulting task set.

\subsubsection{Round-by-Round Construction}

\begin{figure}[t]
\centering
\includegraphics[width=\columnwidth]{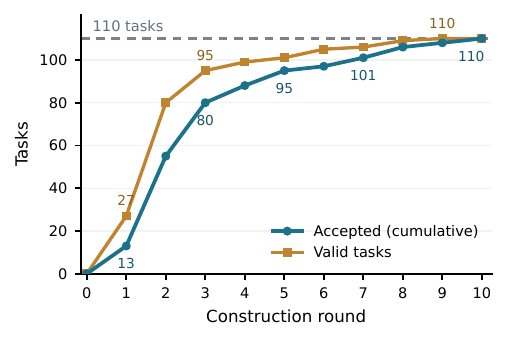}
\caption{Generation progress across 110 tasks. Accepted is cumulative; Valid equals cumulative Accepted plus current-round Harden. The dashed line marks 110 tasks.}
\label{fig:genprogress}
\end{figure}

Figure~\ref{fig:genprogress} tracks construction progress across 110 task lineages, initialized as five modeling and five debugging tasks from each of 11 cleaned financial models. Accept denotes valid challenges at the target difficulty; Harden denotes valid but solvable candidates. Each round\textquotesingle{}s valid task count equals cumulative Accept plus current-round Harden.

The accepted tasks operate on workbooks with a median of eight worksheets (range: 1--115) and a median of 51 scored target cells (range: 3--4,548). These sizes describe the artifact scope; the cases below make the associated reasoning demands concrete through driver selection, period alignment, cash-flow dependencies and the interpretation of business assumptions. The resulting solving trajectories exceed 150k tokens of cumulative context on average in both training and development evaluation (192.5k and 159.3k, respectively; Appendix~\ref{app:trajectories}).

\textbf{The first-round results highlight the need for iterative construction.} Only 13 candidates are accepted and 14 proceed to hardening, giving 27 valid tasks (24.55\pct) and a target-difficulty yield of 11.82\pct. The remaining 83 (75.45\pct) require further work, including 76 that fail pre-solve checks (Table~\ref{tab:roundflow}, Appendix~\ref{app:process}).

\textbf{Most challenges reach the retention criterion within three rounds.} By round three, 95 tasks (86.36\pct) are valid in their current versions and 80 (72.73\pct) have been accepted. Valid task count reaches 110 at round nine, when 108 are accepted; all 110 are accepted by round ten.

\textbf{A minority of tasks account for over half of the construction effort.} The 30 lineages requiring more than three rounds represent 27.27\pct{} of tasks but consume 185/357 (51.82\pct) of model-authored versions. These lineages require an average of 6.17 authored versions per task, compared with 2.15 for the 80 tasks accepted within three rounds, a 2.87-fold difference in construction effort.

\subsubsection{Success-Driven Hardening}

\begin{figure}[t]
\centering
\includegraphics[width=\columnwidth]{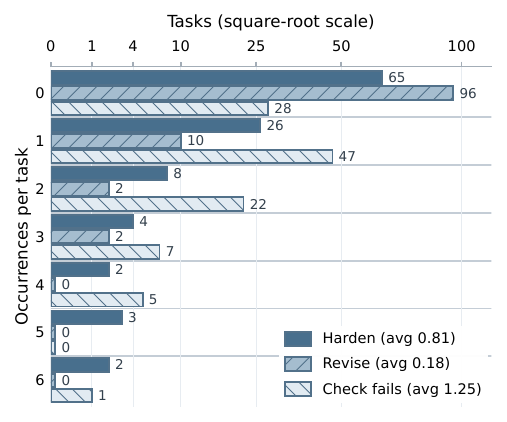}
\caption{Per-task frequencies of Harden, Revise and Check fails across 110 tasks, with a square-root task-count axis. The legend reports average occurrences per task, including zeros; outcomes follow content-round accounting. Short stubs mark zero counts.}
\label{fig:feedbackfreq}
\end{figure}

Across the 110 tasks in Figure~\ref{fig:feedbackfreq}, 65 never trigger hardening, 26 trigger it once and 19 trigger it at least twice, with a maximum of six times. Thus 45 tasks (40.91\pct) undergo 89 hardening steps. Appendix~\ref{app:process} reports the successor outcomes and target-count comparisons underlying the following analysis.

\textbf{Trajectory-guided hardening produces challenges beyond the current solver.} The author revisits the successful trajectory, script and output to identify formula examples, repeated structures or directly reusable computations, then adjusts the next challenge accordingly; Appendix~\ref{app:process} documents this feedback protocol. Of 68 immediate hardened successors with normal solver outcomes, 37 (54.41\pct) are no longer solved successfully, and 33 of these are accepted after author review (Table~\ref{tab:hardenoutcomes}). The remaining four failures are revised. These observed transitions show how successful solves supply actionable feedback for further challenge construction.

\textbf{Hardening increases business-semantic and dependency-reasoning demands.} Among 67 comparable before/after pairs, target count increases in 40, stays unchanged in 23 and decreases in four; eight failed successors have unchanged counts and two have fewer targets (Table~\ref{tab:hardensize}). The five-target example in Figure~\ref{fig:cases}A replaces cash-flow sign/operator faults with driver selection, aggregation scope, period alignment and lease roll-forward errors. Table~\ref{tab:hardenrelations} details all five changes and their original target outcomes: four are corrected, while the next-period CapEx-rate error remains. The CapEx formula remains executable, but selecting the next-period rate violates the intended temporal dependency. Identifying this error requires matching the formula to its business driver and period, illustrating increased semantic and dependency-reasoning demands without adding targets.

\subsubsection{Check- and Review-Driven Task Repair}
\label{sec:repair}

Figure~\ref{fig:feedbackfreq} summarizes the two repair routes: 138 Check fails returns across 82 tasks and 20 review-triggered Revise decisions across 14 tasks. Table~\ref{tab:repairflow} records their immediate successor outcomes; Table~\ref{tab:repairprovenance} links the accepted tasks to their preceding transitions.

\textbf{The two repair routes jointly form a major path to the accepted corpus.} Of 110 accepted tasks, 88 (80.00\pct) encounter at least one Check fails or Revise; eight encounter both. Moreover, 63 tasks (57.27\pct) are accepted in the immediate successor version after a return: 54 after Check fails and nine after Revise (Table~\ref{tab:repairprovenance}). Thus repair participates in most task histories and directly precedes acceptance for over half of the corpus, rather than merely accounting for isolated edits at the end of construction.

\textbf{Check failures expose unmet verification conditions before independent solving.} Of 138 returns, 136 arise during construction/replay checks, one from the unchanged-input negative control and one at the reference gate (Appendix~\ref{app:process}). In the negative-control case, leaving the input unchanged already earns a passing score, exposing a candidate that does not require the intended repair. Such feedback identifies a concrete validation condition to revisit before interpreting solver performance as task difficulty. These checks therefore support task construction through executable feedback, without relying on a solver failure to reveal the problem.

\textbf{Failure review uses solver behavior to uncover task issues after checks pass.} The 130 normal solver failures lead to 20 Revise decisions, of which 19 cite task issues. The reports include six missing-public-assumption cases, four instruction-scope or meaning cases and three structural-target cases (Appendix~\ref{app:process}). In F04-D002, review identifies missing public allocation evidence and prompts explicit category labels and ratios to be restored while all 28 targets and their formulas remain unchanged (Figure~\ref{fig:cases}B, Appendix~\ref{app:cases}). The unchanged formulas show why executable checks alone do not establish that the public business assumptions are sufficient: the solver must know which categories the allocation ratios describe. Revise uses the trajectory and output to diagnose this semantic gap after checks pass, complementing returns tied to unmet executable conditions. Taken together, the loop yields 110 accepted tasks, and 37 of 68 hardened successors with normal solver outcomes are not solved successfully; check and review feedback address task defects.

\subsection{RL Self-Improvement (RQ2)}
\label{sec:learning}

We split the 110 generated tasks into 88 training tasks and 22 held-out within-source development tasks and measure policy performance before and after training.

\subsubsection{Training Procedure}

We perform eight synchronous GRPO-style updates using fresh target-policy interactions on the generated tasks and fixed whole-task rewards. At each development checkpoint, the current policy makes five independent attempts on each of the same 22 tasks. We report avg5 scores by averaging the five attempt scores within each task and then averaging across tasks.

\subsubsection{\revision{Held-Out Within-Source Evaluation}}

\begin{figure}[t]
\centering
\includegraphics[width=\columnwidth]{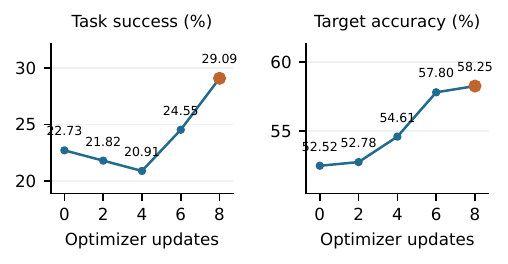}
\caption{Learning progress before RL and through eight updates: avg5 over 22 tasks with five attempts each. Orange marks the update-eight result.}
\label{fig:learning}
\end{figure}

\textbf{Both whole-task success and target accuracy improve over the pretrained reference.} From before training to update eight, avg5 task success rises from 22.73\pct{} to 29.09\pct{}, a gain of 6.36 percentage points or 28.0\pct{} relative, while target accuracy rises from 52.52\pct{} to 58.25\pct{}. Target accuracy increases at every measured checkpoint; task success initially declines before improving at updates six and eight (Figure~\ref{fig:learning}). These metrics capture complementary changes: target accuracy records partial progress at the target level, whereas whole-task success requires every target to be correct.

\textbf{Both task types improve in whole-task success, while modeling drives target-accuracy gains.} From the pretrained reference to update eight, avg5 task success rises from 23.64\pct{} to 30.91\pct{} for modeling and from 21.82\pct{} to 27.27\pct{} for debugging (Table~\ref{tab:rltypes}). Modeling target accuracy increases from 54.18\pct{} to 66.83\pct{}, whereas debugging changes from 50.85\pct{} to 49.66\pct{}. Thus both task types produce more fully correct outputs, while target-level gains concentrate in modeling, showing held-out gains in this single cycle.

\begin{table}[t]
\centering\small\setlength{\tabcolsep}{4pt}
\begin{tabular}{@{}lrrrr@{}}
\toprule
 & \multicolumn{2}{c}{Task success (\%)} & \multicolumn{2}{c}{Target accuracy (\%)}\\
\cmidrule(lr){2-3}\cmidrule(lr){4-5}
Type & Before RL & Step 8 & Before RL & Step 8\\
\midrule
Modeling & 23.64 & 30.91 & 54.18 & 66.83\\
Debugging & 21.82 & 27.27 & 50.85 & 49.66\\
\bottomrule
\end{tabular}
\caption{Task-type results before training and after eight updates. Task success is avg5; target accuracy is macro-averaged.}
\label{tab:rltypes}
\end{table}

\subsubsection{\revision{Generalization of Financial Modeling Capabilities}}
\label{sec:external-transfer}

\begin{figure}[t]
\centering
\includegraphics[width=\columnwidth]{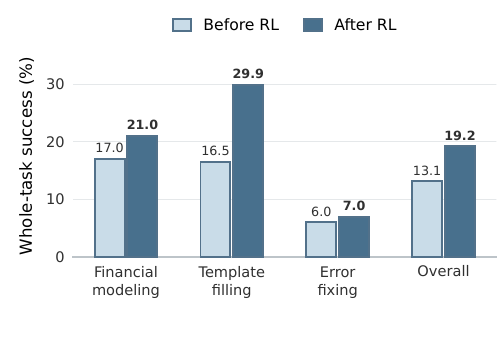}
\caption{\revision{Whole-task success on SpreadsheetBench 2 Financial before and after RL.}}
\label{fig:external-transfer}
\end{figure}

\revision{\textbf{Learning gains persist across task sources and evaluation settings.} Whole-task success on SpreadsheetBench 2 Financial rises from 13.13\pct{} (39/297) to 19.19\pct{} (57/297), a gain of 6.06 percentage points, close to the 6.36-point within-source gain (Figure~\ref{fig:external-transfer}). The overall improvement retains a similar magnitude under a change in task sources and scoring protocol, providing evidence that the learning benefits generalize beyond the construction setting.}

\revision{\textbf{The largest gain occurs in a task category absent from training.} Template-filling success increases by 13.40 percentage points (16/97 to 29/97), although training includes only modeling and debugging. A plausible explanation is that demanding modeling and debugging challenges exercise business-relation understanding, dependency reasoning and spreadsheet operations also needed for template filling. Combined with the comparable within-source and external gains, this pattern supports improvements in underlying financial modeling capabilities shared across tasks, rather than gains confined to particular source workbooks or task formats.}

\subsection{Independent Solvability Check}
\label{sec:validation}

We conduct a small-scale supplementary validation on eight randomly selected tasks generated by \method{}. GPT-6/Codex completes six tasks (6/8), while Qwen/OpenHands completes none (0/8). These six successful independent solutions provide additional evidence that \method{} can generate tasks that challenge the target configuration while remaining solvable.

\section{Related Work}

\paragraph{Recursive self-improvement and agent training.}
STOP \citep{stop} and the Darwin G\"odel Machine \citep{dgm} improve agent programs, while Socratic learning \citep{socraticlearning} frames agent outputs as future learning inputs. Absolute Zero \citep{absolutezero}, MAE \citep{mae} and STRETCH \citep{stretch} couple task generation and solving within one model. R-Zero \citep{rzero} uses separate challenger and solver models; CURE \citep{cure} co-evolves coding and unit-test generation through execution rewards. R-Quest \citep{rquest} studies validity and novelty feedback for sustained question generation. FinEvo examines feedback-guided experience construction and learning in financial modeling.

\paragraph{Generating tasks from existing artifacts.}
SWE-smith \citep{swesmith} introduces bugs into repositories to create repair tasks; Workbook Time Machine \citep{wtm} removes derived workbook artifacts to create reconstruction tasks. EnvHarness \citep{envharness} reshapes existing environments through programmable components while preserving their verifiers; its EnvRigger diagnoses trajectories and validates candidates with fresh rollouts. FinEvo uses financial workbooks as references, with successful solves guiding hardening and failure review addressing task defects before learning.

\paragraph{Spreadsheet tasks and agent training.}
SpreadsheetBench \citep{spreadsheetbench} evaluates forum-derived spreadsheet operations; SpreadsheetBench 2 \citep{spreadsheetbench2} covers end-to-end workflows, including expert-authored financial modeling and debugging. Spreadsheet-RL \citep{spreadsheetrl} uses strong coding agents to construct workbook pairs and trains spreadsheet policies with outcome-based GRPO. FinEvo instead investigates whether the target model's own solving feedback can guide challenge construction and whether learning from its own interactions improves held-out financial task performance.

\section{Conclusion}

\revision{\method{} evaluates one complete construction--learning cycle for long-horizon financial modeling. Existing financial capabilities support challenge evolution, while learning from the resulting experience yields comparable gains across task sources and evaluation settings and improves a task category absent from training. These results support the value of self-generated challenges for strengthening underlying financial modeling capabilities shared across tasks, extending the empirical exploration of RSI to a high-value professional domain. Broader domain transfer and repeated-generation behavior remain to be tested.}

\section*{Limitations}

\revision{\textbf{Task scope and transfer.} Task construction and external evaluation are limited to financial spreadsheets; transfer to other professional domains remains unverified. The shared-capability explanation and the contribution of challenge hardening to transfer require controlled ablations.}

\textbf{Model scale.} Our experiments use Qwen3.6-27B, and the RSI loop has yet to be evaluated with larger-parameter models. Whether the observed task-construction and learning benefits persist as model scale increases remains to be established.

\section*{Ethical Considerations and Data Access}

The financial models originated in nonpublic materials. We anonymize identities and synthesize business values during preparation; these transformations do not establish redistribution rights, and we do not release the workbooks as a public benchmark. No original workbooks, company/provider identities or financial values appear in the anonymous numerical reproduction bundle. Any later task release requires the relevant rights and privacy checks. Complete training reproduction requires authorized model and task access; CPU statistical reproduction is a different level of access.

The deterministic scorer does not certify financial suitability, and self-review can miss semantic defects. Passing tasks should not be used as financial advice. No human participants or new personal annotations were collected. We used AI assistance for implementation, analysis, writing and internal simulated review, and remain responsible for the evidence and submission declarations.

\bibliography{references,rsi-references,intro_2-references}

@misc{mlmaster,
  title = {{ML-Master}: Towards {AI-for-AI} via Integration of Exploration and Reasoning},
  author = {Liu, Zexi and Cai, Yuzhu and Zhu, Xinyu and Zheng, Yujie and Chen, Runkun and Wen, Ying and Wang, Yanfeng and E, Weinan and Chen, Siheng},
  year = {2025},
  eprint = {2506.16499},
  archivePrefix = {arXiv},
  url = {https://arxiv.org/abs/2506.16499}
}

@misc{frontisma1,
  title = {{Frontis-MA1}: Training an {AI4AI} Model towards Recursive Self-Improvement in Machine Learning Engineering},
  author = {Yang, Junlin and Jiang, Che and Fu, Yu and Luo, Tianwei and Ren, Can and Wang, Weizhi and Zhao, Kaikai and Liu, Hongyi and Zuo, Yuxin and Wang, Yuru and Fan, Yuchen and Tian, Kai and Yuan, Zhenzhao and Lin, Xiaojian and Sheng, Li and Qiang, Rushi and Jia, Guoli and Lv, Xingtai and Hua, Ermo and Lei, Dianqiao and Sun, Youbang and Ding, Ning and Zhou, Bowen and Zhang, Kaiyan},
  year = {2026},
  eprint = {2607.28568},
  archivePrefix = {arXiv},
  url = {https://arxiv.org/abs/2607.28568}
}

@misc{anthropicfinance,
  author = {{Anthropic}},
  title = {{Claude for Financial Services}},
  year = {2025},
  month = jul,
  howpublished = {Official product announcement, July 15},
  url = {https://www.anthropic.com/news/claude-for-financial-services},
  note = {Accessed November 13, 2026}
}

@techreport{icaewmodellingcode,
  author = {{ICAEW}},
  title = {Financial Modelling Code},
  institution = {Institute of Chartered Accountants in England and Wales},
  year = {2024},
  url = {https://www.icaew.com/-/media/corporate/files/technical/technology/excel/financial-modelling-code.ashx},
  note = {See Model definition and purpose, p. 3, and Review and test your models, p. 14}
}

@inproceedings{graves2017curriculum,
  author = {Graves, Alex and Bellemare, Marc G. and Menick, Jacob and Munos, R{\'e}mi and Kavukcuoglu, Koray},
  title = {Automated Curriculum Learning for Neural Networks},
  booktitle = {Proceedings of the 34th International Conference on Machine Learning},
  series = {Proceedings of Machine Learning Research},
  volume = {70},
  pages = {1311--1320},
  year = {2017},
  publisher = {PMLR},
  url = {https://proceedings.mlr.press/v70/graves17a.html}
}

@inproceedings{sukhbaatar2018selfplay,
  author = {Sukhbaatar, Sainbayar and Lin, Zeming and Kostrikov, Ilya and Synnaeve, Gabriel and Szlam, Arthur and Fergus, Rob},
  title = {Intrinsic Motivation and Automatic Curricula via Asymmetric Self-Play},
  booktitle = {International Conference on Learning Representations},
  year = {2018},
  url = {https://arxiv.org/abs/1703.05407}
}

@misc{envharness,
  title = {{EnvHarness}: Awakening Static Worlds for Agent Learning},
  author = {Huang, Chengsong and Wang, Zifeng and Han, Rujun and Yan, Jun and Chen, Yanfei and CuiZhu, Zoey and Jiang, Ke and Xia, Peng and Yu, Han and Zhuang, Yufan and Ming, Yifei and Pan, Jiaqi and Mishra, Bhavana Dalvi and Huang, Jiaxin and Gokturk, Burak and Pfister, Tomas and Lee, Chen-Yu},
  year = {2026},
  eprint = {2608.19880},
  archivePrefix = {arXiv},
  url = {https://arxiv.org/abs/2608.19880}
}

@misc{spreadsheetbench,
  title = {{SpreadsheetBench: Towards Challenging Real World Spreadsheet Manipulation}},
  author = {Ma, Zeyao and Zhang, Bohan and Zhang, Jing and Yu, Jifan and Zhang, Xiaokang and Zhang, Xiaohan and Luo, Sijia and Wang, Xi and Tang, Jie},
  year = {2024},
  eprint = {2406.14991},
  archivePrefix = {arXiv},
  url = {https://arxiv.org/abs/2406.14991}
}

@misc{spreadsheetbench2,
  title = {{SpreadsheetBench 2: Evaluating Agents on End-to-End Business Spreadsheet Workflows}},
  author = {Zhu, Jian and Zhang, Yuzheng and Ma, Zeyao and Zhang, Bohan and Schoepf, Armin and Woloch, Daniel and Wang, Peter Yiliu and Yang, Guangyu Robert and Jacob, Samuel and Nagisetty, Siddharth and Chundru, Abhiram and Lin, Jean and Mateega, Spencer and Zhang, Jing},
  year = {2026},
  eprint = {2606.29955},
  archivePrefix = {arXiv},
  url = {https://arxiv.org/abs/2606.29955}
}

@misc{cfainstitute,
  author = {{CFA Institute}},
  title = {Financial Modeling},
  year = {n.d.},
  howpublished = {CFA Program Practical Skills Module},
  url = {https://www.cfainstitute.org/programs/cfa-program/candidate-resources/practical-skills-modules/financial-modelling},
  note = {Accessed November 11, 2026}
}

@misc{frontierfinance,
  title = {{FrontierFinance}: A Long-Horizon Computer-Use Benchmark of Real-World Financial Tasks},
  author = {Krumdick, Michael and Reddy, Varshini and Chaudhary, Shivani and Day, William and Ahmed, Maarij and Haqqi, Hayan and Fahim, Muhammad Ahsen and Amjad, Hanzallah and Orakzai, Ahmad and Gul, Aqsa and Tanner, Chris},
  year = {2026},
  eprint = {2604.05912},
  archivePrefix = {arXiv},
  url = {https://arxiv.org/abs/2604.05912}
}

@misc{kimiK31Pro,
  author = {{Moonshot AI}},
  title = {Introducing {Kimi} K3.1 Pro},
  year = {n.d.},
  howpublished = {Kimi Technical Blog},
  url = {https://www.kimi.ai/blog/kimi-k3-1-pro},
  note = {Accessed November 11, 2026}
}

@misc{spreadsheetrl,
  title = {{Spreadsheet-RL: Advancing Large Language Model Agents on Realistic Spreadsheet Tasks via Reinforcement Learning}},
  author = {Chi, Banghao and Xie, Yining and Wu, Mingyuan and Yang, Jingcheng and Jiang, Jize and Li, Zhaoheng and Qian, Shengyi and Zhang, Minjia and Nahrstedt, Klara and Hou, Rui and Fan, Xiangjun and Yu, Hanchao},
  year = {2026},
  eprint = {2605.22642},
  archivePrefix = {arXiv},
  url = {https://arxiv.org/abs/2605.22642}
}

@misc{wtm,
  title = {{Back to the Future: A workbook time machine for spread sheet creation benchmarks}},
  author = {Uniyal, Mansi and Singh, Agamdeep and Singha, Ananya and Gupta, Priyanshu and Singh, Mukul and Verbruggen, Gust and Le, Vu and Gulwani, Sumit},
  year = {2026},
  eprint = {2608.07873},
  archivePrefix = {arXiv},
  url = {https://arxiv.org/abs/2608.07873}
}

@misc{deepseekmath,
  title = {{DeepSeekMath: Pushing the Limits of Mathematical Reasoning in Open Language Models}},
  author = {Shao, Zhihong and Wang, Peiyi and Zhu, Qihao and Xu, Runxin and Song, Junxiao and Bi, Xiao and Zhang, Haowei and Zhang, Mingchuan and Li, Y. K. and Wu, Y. and Guo, Daya},
  year = {2024},
  eprint = {2402.03300},
  archivePrefix = {arXiv},
  url = {https://arxiv.org/abs/2402.03300}
}

@misc{openhands,
  title = {{OpenHands: An Open Platform for AI Software Developers as Generalist Agents}},
  author = {Wang, Xingyao and Li, Boxuan and Song, Yufan and Xu, Frank F. and Tang, Xiangru and Zhuge, Mingchen and Pan, Jiayi and Song, Yueqi and Li, Bowen and Singh, Jaskirat and Tran, Hoang H. and Li, Fuqiang and Ma, Ren and Zheng, Mingzhang and Qian, Bill and Shao, Yanjun and Muennighoff, Niklas and Zhang, Yizhe and Hui, Binyuan and Lin, Junyang and Brennan, Robert and Peng, Hao and Ji, Heng and Neubig, Graham},
  year = {2024},
  eprint = {2407.16741},
  archivePrefix = {arXiv},
  url = {https://arxiv.org/abs/2407.16741}
}

@misc{modaldojo,
  author = {{Modal}},
  title = {Training multi-turn coding agents},
  year = {2026},
  howpublished = {Modal Dojo tutorial},
  url = {https://dojo.modal.dev/tutorials/coding_agent},
  note = {Accessed October 16, 2026; local recipe file hashes retained}
}

@misc{slime,
  author = {{Slime Contributors}},
  title = {Slime: An {LLM} Post-Training Framework},
  year = {2026},
  url = {https://github.com/modal-projects/slime/tree/3585d4a7eb1a5c108810238b47c37d3107d0a2ba},
  note = {Modal Dojo fork; source revision 3585d4a7eb1a5c108810238b47c37d3107d0a2ba}
}

@misc{qwen36,
  author = {{Qwen Team}},
  title = {{Qwen3.6-27B} Model Card},
  year = {2026},
  url = {https://huggingface.co/Qwen/Qwen3.6-27B},
  note = {Accessed October 16, 2026}
}

@misc{absolutezero,
  title = {Absolute Zero: Reinforced Self-play Reasoning with Zero Data},
  author = {Zhao, Andrew and Wu, Yiran and Yue, Yang and Wu, Tong and Xu, Quentin and Lin, Matthieu and Wang, Shenzhi and Wu, Qingyun and Zheng, Zilong and Huang, Gao},
  year = {2025},
  eprint = {2505.03335},
  archivePrefix = {arXiv},
  url = {https://arxiv.org/abs/2505.03335}
}

@misc{rzero,
  title = {{R-Zero}: Self-Evolving Reasoning {LLM} from Zero Data},
  author = {Huang, Chengsong and Yu, Wenhao and Wang, Xiaoyang and Zhang, Hongming and Li, Zongxia and Li, Ruosen and Huang, Jiaxin and Mi, Haitao and Yu, Dong},
  year = {2025},
  eprint = {2508.05004},
  archivePrefix = {arXiv},
  url = {https://arxiv.org/abs/2508.05004}
}

@inproceedings{mae,
  title = {Multi-Agent Evolve: {LLM} Self-Improve through Co-evolution},
  author = {Chen, Yixing and Wang, Yiding and Zhu, Siqi and Xu, Weixian and Yu, Haofei and Feng, Tao and Zhang, Muhan and Patwary, Mostofa and You, Jiaxuan},
  booktitle = {Proceedings of the 2026 Conference on Empirical Methods in Natural Language Processing},
  pages = {39563--39593},
  address = {Budapest, Hungary},
  publisher = {Association for Computational Linguistics},
  year = {2026},
  doi = {10.18653/2026.emnlp-main.1734},
  url = {https://aclanthology.org/2026.emnlp-main.1734/}
}

@misc{swesmith,
  title = {{SWE-smith}: Scaling Data for Software Engineering Agents},
  author = {Yang, John and Lieret, Kilian and Jimenez, Carlos E. and Wettig, Alexander and Khandpur, Kabir and Zhang, Yanzhe and Hui, Binyuan and Press, Ofir and Schmidt, Ludwig and Yang, Diyi},
  year = {2025},
  eprint = {2504.21798},
  archivePrefix = {arXiv},
  url = {https://arxiv.org/abs/2504.21798}
}

@inproceedings{stretch,
  title = {{STRETCH} the Boundaries: A Unified Self-Taught Framework for Progressive {LLM} Evolution},
  author = {Yu, Yajie and Lee, Mark and Feng, Yue},
  booktitle = {Findings of the Association for Computational Linguistics: EMNLP 2026},
  year = {2026},
  pages = {24732--24747},
  doi = {10.18653/2026.findings-emnlp.1168},
  url = {https://aclanthology.org/2026.findings-emnlp.1168/}
}

@misc{rquest,
  title = {Questioning the Questions: Sustaining Self-Evolution in Reasoning Models},
  author = {Li, Jinyuan and Huang, Chengsong and Huang, Langlin and Cai, Donghong and Gao, Shiping and Yang, Yuyi and Huang, Jiaxin},
  year = {2026},
  note = {arXiv preprint arXiv:2610.04299},
  eprint = {2610.04299},
  archivePrefix = {arXiv},
  url = {https://arxiv.org/abs/2610.04299}
}

@inproceedings{stop,
  title = {Self-Taught Optimizer ({STOP}): Recursively Self-Improving Code Generation},
  author = {Zelikman, Eric and Lorch, Eliana and Mackey, Lester and Kalai, Adam Tauman},
  booktitle = {First Conference on Language Modeling},
  year = {2024},
  url = {https://arxiv.org/abs/2310.02304}
}

@misc{dgm,
  title = {Darwin {Gödel} Machine: Open-Ended Evolution of Self-Improving Agents},
  author = {Zhang, Jenny and Hu, Shengran and Lu, Cong and Lange, Robert and Clune, Jeff},
  year = {2025},
  eprint = {2505.22954},
  archivePrefix = {arXiv},
  url = {https://arxiv.org/abs/2505.22954}
}

@misc{socraticlearning,
  title = {Boundless Socratic Learning with Language Games},
  author = {Schaul, Tom},
  year = {2024},
  eprint = {2411.16905},
  archivePrefix = {arXiv},
  url = {https://arxiv.org/abs/2411.16905}
}

@inproceedings{cure,
  title = {Co-Evolving {LLM} Coder and Unit Tester via Reinforcement Learning},
  author = {Wang, Yinjie and Yang, Ling and Tian, Ye and Shen, Ke and Wang, Mengdi},
  booktitle = {Advances in Neural Information Processing Systems},
  volume = {38},
  year = {2025},
  doi = {10.52202/085713-4809},
  url = {https://proceedings.neurips.cc/paper_files/paper/2025/hash/d38653cdaa8e992549e1e9e1621610d7-Abstract-Conference.html}
}

@misc{coworkos,
  title = {{CoworkOS}: Scaling Agent Coworkers for Real-world Automation},
  author = {Jin, Jiajie and Dong, Guanting and Pan, Chengjun and Deng, Xin and Sheng, Zhuohui and Wang, Guoqiang and Gui, Tao and Zhu, Yutao and Chen, Yujia and Yi, Shihong and Hu, Binze and Yang, Zhixiong and Sun, Zhichao and Wang, Rui and Han, Han and Ma, Mingyu and Huang, Junhao and Wang, Xiaotao and Jiang, Huicheng and Hu, Yuyang and Yang, Xinyu and Xu, Wanghan and Yang, Yajie and Wang, Zhecan and Guo, Honglin and Xing, Xiaoyu and Zhang, Xiaoqian and Zhu, Yulun and Li, Xiaoxi and Lin, Jiahang and Liu, Shaofan and Dou, Shihan and Xi, Zhiheng and Zhang, Qi and Jiang, Yu-Gang and Ng, See-Kiong and Zhang, Yang and Chua, Tat-Seng and Wen, Ji-Rong and Dou, Zhicheng},
  year = {2026},
  eprint = {2611.03178},
  archivePrefix = {arXiv},
  url = {https://arxiv.org/abs/2611.03178}
}

\appendix
\section{Process Counts and Review Coding}
\label{app:process}

\paragraph{Trajectory-guided hardening feedback.}
The archived hardening prompt supplies the prior successful trajectory, script, output and scores to the same-model author. It asks the author to identify exploitable formula examples, repeated structures, independent computation modules or directly reusable relations, and to state the next hardening mechanism before generating a new candidate. It permits changes to business derivations, information distribution, cross-sheet dependencies and boundary conditions, while requiring sufficient public evidence. This documents the feedback supplied to the author; the observed successor outcomes do not isolate its effect against a no-trajectory control.

\begin{table}[t]
\centering\small\setlength{\tabcolsep}{3pt}
\begin{tabular}{@{}lrr@{}}
\toprule
Immediate successor outcome & Count & \% of 68\\
\midrule
Solved; further hardening & 31 & 45.59\\
Failed; accepted after review & 33 & 48.53\\
Failed; revised after review & 4 & 5.88\\
\midrule
Normal solver outcomes & 68 & 100.00\\
No normal solver outcome & 21 & --\\
All hardened successors & 89 & --\\
\bottomrule
\end{tabular}
\caption{Immediate outcomes of 89 hardened versions. Percentages use the 68 normal solver outcomes; the other 21 do not enter this denominator.}
\label{tab:hardenoutcomes}
\end{table}

\begin{table}[t]
\centering\small
\begin{tabular}{@{}lrrr@{}}
\toprule
Target-count change & Solved & Failed & Total\\
\midrule
Increase & 14 & 26 & 40\\
Unchanged & 15 & 8 & 23\\
Decrease & 2 & 2 & 4\\
\midrule
Total & 31 & 36 & 67\\
\bottomrule
\end{tabular}
\caption{Hardening outcomes by target-count change for 67 comparable parent/child pairs. One of the 68 normal successor trials lacks comparable target counts.}
\label{tab:hardensize}
\end{table}

\paragraph{Check- and review-driven repair transitions.}
Table~\ref{tab:repairflow} counts one immediate content-round successor per return, including repeated events within the same task. Eight tasks encounter both routes. Check fails comprises 136 construction/replay returns, one negative-control rejection and one reference-gate return. These stages are distinct from normal solver failures and do not establish that every returned candidate is semantically unsolvable. The successor counts describe observed revisions, not repair effectiveness against an unrepaired control. Table~\ref{tab:repairprovenance} counts each final accepted task once by the disposition immediately before its accepted content version; initial acceptance is separate. The 88 ever-returned tasks are the union of the 82 Check fails and 14 Revise lineages, with eight shared lineages.

\begin{table*}[t]
\centering\small
\begin{tabular}{@{}lrrrrrr@{}}
\toprule
Trigger & Tasks & Events & Next Check fails & Next Harden & Next Revise & Next Accept\\
\midrule
Check fails & 82 & 138 & 38 & 38 & 8 & 54\\
Revise & 14 & 20 & 4 & 6 & 1 & 9\\
\bottomrule
\end{tabular}
\caption{Immediate outcomes after Check fails or Revise. The four successor-outcome columns sum to Events; Tasks counts distinct lineages within each row.}
\label{tab:repairflow}
\end{table*}

\begin{table}[t]
\centering\small
\begin{tabular}{@{}lr@{}}
\toprule
Preceding transition & Accepted tasks\\
\midrule
Initial candidate & 13\\
Harden & 34\\
Check fails & 54\\
Revise & 9\\
\midrule
Total & 110\\
\bottomrule
\end{tabular}
\caption{Final accepted tasks by preceding transition. Each task appears once. Across complete histories, 88 tasks encounter at least one Check fails or Revise.}
\label{tab:repairprovenance}
\end{table}

\paragraph{Round-wise candidate outcomes.}
Table~\ref{tab:roundflow} provides the per-round dispositions underlying the cumulative progress in Figure~\ref{fig:genprogress}.

\begin{table}[t]
\centering\small\setlength{\tabcolsep}{3pt}
\begin{tabular}{@{}rrrrrrr@{}}
\toprule
Round & In & Accept & Harden & Revise & Check fails & Next\\
\midrule
1 & 110 & 13 & 14 & 7 & 76 & 97\\
2 & 97 & 42 & 25 & 0 & 30 & 55\\
3 & 55 & 25 & 15 & 6 & 9 & 30\\
4 & 30 & 8 & 11 & 3 & 8 & 22\\
5 & 22 & 7 & 6 & 2 & 7 & 15\\
6 & 15 & 2 & 8 & 0 & 5 & 13\\
7 & 13 & 4 & 5 & 1 & 3 & 9\\
8 & 9 & 5 & 3 & 1 & 0 & 4\\
9 & 4 & 2 & 2 & 0 & 0 & 2\\
10 & 2 & 2 & 0 & 0 & 0 & 0\\
\midrule
Total & 357 & 110 & 89 & 20 & 138 & --\\
\bottomrule
\end{tabular}
\caption{Final outcomes by model-authored task-content round, mapped to Figure~\ref{fig:workflow}B. Check fails follows \emph{Check fails: reconstruct}; Revise corresponds to \emph{Repair task} after failure review. Successful solves trigger Harden; acceptance follows failure review. Next equals Harden + Revise + Check fails. Returned candidates undergo construction, checks and solving again; accepted tasks exit. The last row reports cumulative counts.}
\label{tab:roundflow}
\end{table}

\begin{figure*}[t]
\centering
\includegraphics[width=\textwidth]{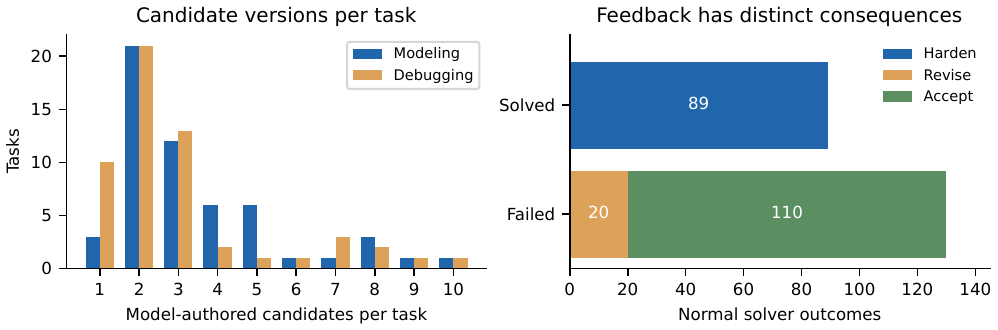}
\caption{Candidate and feedback counts. Left: model-authored versions per task, excluding one mechanical packaging revision. Right: 219 normal solver outcomes lead to 89 hardenings, 20 review revisions and 110 acceptances. These are selected construction histories, not independent benchmark rates or a generation ablation.}
\label{fig:generation}
\end{figure*}

\begin{table*}[t]
\centering\small
\begin{tabular}{lrrrrrr}
\toprule
Source & M & D & Reject & Harden & Revise & Missing\\
\midrule
F01 & 11 & 13 & 10 & 4 & 0 & 0 \\
F02 & 18 & 12 & 14 & 5 & 1 & 1 \\
F03 & 14 & 25 & 13 & 14 & 1 & 1 \\
F04 & 13 & 9 & 9 & 1 & 2 & 0 \\
F05 & 11 & 10 & 6 & 3 & 2 & 0 \\
F06 & 20 & 15 & 14 & 11 & 0 & 2 \\
F07 & 22 & 19 & 14 & 16 & 1 & 1 \\
F08 & 23 & 14 & 12 & 11 & 3 & 1 \\
F09 & 16 & 13 & 11 & 2 & 6 & 2 \\
F10 & 25 & 27 & 23 & 16 & 4 & 0 \\
F11 & 18 & 9 & 11 & 6 & 0 & 0
\\
\midrule
Total & 191 & 166 & 137 & 89 & 20 & 8\\
\bottomrule
\end{tabular}
\caption{Anonymous source-level histories. Each source starts five tasks per type. M/D count authored modeling/debugging versions, excluding the mechanical version. Reject counts replay/package rejections; Harden and Revise count transitions; Missing counts final selected attempts without a readable target report. These columns describe different stages and are not additive.}
\label{tab:sourceprocess}
\end{table*}

\paragraph{Failure-review issue coding.}
Single-analyst coding of the 20 revised reviews assigns one primary reported issue per version: six missing public assumptions, four instruction-scope or meaning issues, three structural-target issues, five reference/build/check or recalculation issues, one claimed budget/scope issue and one solver-attributed revision. For example, reports identify deleted growth assumptions, ambiguous amount-versus-share units and intentional blanks treated as targets. The distribution shows what the author attempted to repair: it often revisited the problem specification or answer contract rather than merely reducing difficulty. These are coded author reports, with repeated versions retained, not independently verified defect rates.

Coding uses each review's summary and root causes. The anonymous coding file binds each decision to a task version and original evidence hash. Mixed cases retain notes; repeated versions remain separate. This single-analyst description of author reports has no inter-rater reliability estimate and does not independently validate the claimed defects or successful repairs. Check fails and Revise are distinguished by their trigger and available evidence, not mutually exclusive defect types: review can also report reference or construction issues.

For example, the budget/scope review asserts that several thousand targets are infeasible within an hour, an assertion we do not verify. Another proposes changing recalculation code, but the manifest/hash checks confirm that the verifier and scoring contract remain fixed. We therefore distinguish review diagnoses from independently verified changes.

For runtime scope, we inspect the retained solver script for every attempt. Of 219 normally classified trials, 192 configure no explicit condenser and 27 configure the same-model 80-event/60,000-token condenser. The 64 abnormal trials include 51 retained scripts without it and 13 missing scripts, whose policy we leave unknown. Final selected trials comprise 97 without and 13 with it. Anonymous task/version/retry rows and runtime signatures are included in the numerical output; scripts are hashed in the private evidence index. A configured condenser does not prove it activated during a given attempt.

\paragraph{Execution accounting and task-quality diagnostics.}

The candidate counts distinguish 357 model-authored versions from one mechanical packaging version. The 110 tasks constitute the eventual accepted corpus, not an initial-budget yield estimate.

The immediate repair outcomes in Section~\ref{sec:repair} count one child per revised review, including repeated revisions within a lineage. They are distinct from the final accepted-task counts.

All 110 accepted packages retain Oracle reward one and both negative-control rewards zero. Their verifier hashes and scoring-contract hashes are unchanged across tasks. The numerical rule therefore remains fixed when evaluating hardened successors. The single unchanged-input pass is rejected as a candidate defect, rather than counted as solver success on a meaningful challenge.

Acceptance still has a limited meaning. The final construction trials include 102 readable target reports and eight missing or unreadable outputs; 14 trials reach the agent time limit. Timeout and submission are overlapping categories. A reviewed zero with no workbook does not establish a financial reasoning failure. The campaign also retains 64 abnormal solver attempts and 22 abnormal Oracle/control attempts. These execution failures are not additional normally failed financial problems. The complete final corpus is pipeline-accepted; it is not a claim that all 110 tasks have been independently certified as unambiguous financial challenges.

\section{Longitudinal Evidence for Challenge Evolution}
\label{app:cases}

These cases are selected to explain distinct design responses, not to estimate their frequency or certify the whole corpus. The case extraction reads original workbooks, candidate records and solver outputs, exports only booleans/counts and binds each file by hash. It performs no LibreOffice run, new scoring or model invocation. Original rewards remain unchanged.

\paragraph{Restore an identifiable allocation.}
Section~\ref{sec:repair} explains the allocation ambiguity and its revision. Direct inspection confirms that v2 adds public category labels and the two reference ratios in an initially empty assumption area; these agree with the reference multipliers. All 28 target positions and their input/reference formulas remain unchanged. Both solver attempts fail all targets, with no timeout and the same configured absence of a condenser. The author revises v1 and accepts v2. These retained checks support the specific public-evidence change, without independently auditing every business relation.

\paragraph{Change the mechanism without adding targets.}
Table~\ref{tab:casehistory} preserves all seven candidates in F03-D001, including the unsuccessful checks. v6 has five cash-flow sign/operator faults. v7 changes the fault locations and families while retaining five targets and exactly the same reference financial-sheet content. The solver scripts are byte-identical and neither configures a condenser. The v7 input uses $\mathrm{CapEx}_t=\mathrm{Revenue}_t\times\mathrm{rate}_{t+1}$ where the reference and adjacent-period pattern use their own current-period rates. Its saved output retains this erroneous formula, and the original report identifies it as the only wrong target. The review's listed five relations correspond to v7. These inspections explain what changed and what the solver missed, but stochastic execution, different target identities and possible unexamined defects prevent a causal estimate of hardening effectiveness.

\begin{table*}[t]
\centering\small
\begin{tabular}{@{}p{.26\textwidth}p{.48\textwidth}p{.16\textwidth}@{}}
\toprule
Business dependency & Reference-to-input change & Target outcome\\
\midrule
Receivables driver & $\mathrm{DSO}_t\to\mathrm{DIO}_t$ & Correct\\
Contract-balance scope & $\Delta(C+N)\to\Delta C$ & Correct\\
CapEx period & $R_t c_t\to R_t c_{t+1}$ & Missed\\
Accrued-liability driver & Accrual rate replaced by prepaid rate & Correct\\
Lease roll-forward & New lease additions omitted & Correct\\
\bottomrule
\end{tabular}
\caption{Five business-dependency errors in F03-D001/v7 and their original solver outcomes. The preceding v6 has five sign/operator errors and is fully solved.}
\label{tab:hardenrelations}
\end{table*}

In Table~\ref{tab:hardenrelations}, DSO/DIO denote sales/inventory days; $C$ and $N$ are current/noncurrent contract balances; $R$ is revenue and $c$ the CapEx rate. The symbolic changes and outcomes come from retained reference/input workbooks and original target reports. They describe this selected case, not corpus-wide frequencies of hardening mechanisms.

\begin{table}[t]
\centering\small
\begin{tabular}{rllrr}
\toprule
Version & Transition & Replay & Targets & Correct\\
\midrule
1 & Construct & Reject & -- & --\\
2 & Construct & Pass & 5 & 5\\
3 & Harden & Pass & 7 & 7\\
4 & Harden & Reject & -- & --\\
5 & Construct & Reject & -- & --\\
6 & Construct & Pass & 5 & 5\\
7 & Harden & Pass & 5 & 4\\
\bottomrule
\end{tabular}
\caption{Complete candidate history for selected lineage F03-D001. Counts describe original normal solver trials; a check rejection has no solver result. Retry attempts are not extra versions.}
\label{tab:casehistory}
\end{table}

\paragraph{Design implications.}

\paragraph{Check what the constraints identify.}

Preserving a total does not preserve an identifiable allocation. A correct reference and a balance check can coexist with insufficient evidence for individual categories. Review should trace the public assumptions that support each requested calculation.

\paragraph{Describe the changed inference.}

Record the business relationships, dependencies and fault mechanisms changed by hardening alongside target count. The equal-count case explains an observed solved-to-failed transition without attributing it to size. Conversely, the ratio repair improves public evidence without changing the recorded zero reward.

\paragraph{Bind review to the candidate.}

Business claims in a review should be linked to the frozen candidate's actual targets and dependencies, with that correspondence checked explicitly.

\paragraph{Track learning eligibility separately.}

Acceptance records a reviewed failed solve; successful demonstrations and reward variation describe other conditions needed by learning algorithms. Maintaining these records separately can identify where additional sampling, different feedback or a different objective may be needed.

\clearpage

\section{Solver-Trajectory Statistics}

\label{app:trajectories}

\paragraph{Scope.}

We analyze all 2,048 trajectories consumed by the eight reported RL updates, covering all 88 training tasks, and 550 selected development trajectories from checkpoints 0, 2, 4, 6 and 8 (22 tasks, five attempts per checkpoint). Successful, failed and timed-out attempts remain in the population. Statistics are trajectory-weighted and pool repeated tasks and policies.

\paragraph{Context and interaction accounting.}

Cumulative context is the sum of peak prompt-plus-completion token counts over compression-separated solver contexts. A growing request prefix is counted once within each context, rather than summed again at every model call. Retained history and summaries are counted in the contexts where they are supplied; summary-generation calls are excluded. Training counts come from exact serialized token chains and loss masks; evaluation counts come from recorded solver calls and explicit compression boundaries.

A solver step is a completed model response, including responses without an environment action. All tool calls count recorded invocation requests. Environment-tool calls count terminal and file-editor invocations only, excluding thinking, completion and task-tracker tools. A terminal invocation counts once even when its command performs multiple operations. Agent time is the original agent-execution duration. Quantiles use linear interpolation.

Context, output, step and time measurements cover the full population. Tool counts use 2,024 training and 543 evaluation trajectories with retained structured action traces; missing action traces are not assigned zero. Six evaluation trajectories contain unfinished solver requests: known prompt tokens are retained, but unrecorded output tokens and incomplete responses are not imputed, so their token volumes are observed lower bounds.

\paragraph{Context and interaction workloads.}

Table~\ref{tab:trajectorysummary} shows median cumulative contexts of 162.5k tokens in training and 147.1k in evaluation, with respective P90 values of 346.4k and 219.9k. Median solver lengths are 47 and 43 steps, and median environment-tool counts are 41 and 38. Multiple contexts occur in 22.90\pct{} of training and 7.45\pct{} of evaluation trajectories. The largest cumulative context is 1,020.6k tokens across seven segments, not a single model input.

\begin{table*}[t]

\centering\small

\setlength{\tabcolsep}{4pt}

\begin{tabular}{@{}lrrrrrrrr@{}}
\toprule
 & \multicolumn{4}{c}{Training} & \multicolumn{4}{c}{Development evaluation}\\
Metric & Mean & Median & P90 & Max & Mean & Median & P90 & Max\\
\midrule
Cumulative context (k tokens) & 192.5 & 162.5 & 346.4 & 1020.6 & 159.3 & 147.1 & 219.9 & 440.3\\
Peak segment (k tokens) & 155.1 & 153.9 & 222.6 & 244.4 & 148.4 & 147.1 & 218.9 & 239.0\\
Solver output (k tokens) & 46.8 & 41.9 & 79.9 & 136.6 & 44.2 & 37.9 & 83.3 & 142.3\\
Solver steps & 48.2 & 47.0 & 75.0 & 75.0 & 45.2 & 43.0 & 74.0 & 75.0\\
All tool calls & 47.9 & 47.0 & 75.0 & 75.0 & 44.6 & 43.0 & 74.0 & 75.0\\
Environment tool calls & 42.7 & 41.0 & 70.0 & 74.0 & 39.8 & 38.0 & 67.0 & 74.0\\
Context segments & 1.3 & 1.0 & 2.0 & 7.0 & 1.1 & 1.0 & 1.0 & 2.0\\
Agent time (seconds) & 752.6 & 686.7 & 1231.9 & 1800.1 & 687.5 & 614.6 & 1205.6 & 1800.1\\
\bottomrule
\end{tabular}

\caption{Per-trajectory context and interaction distributions. Token quantities are in thousands; cumulative context sums compression-separated segments. Tool-call coverage is reported in the text.}

\label{tab:trajectorysummary}

\end{table*}

\paragraph{Task-type breakdown.}

Table~\ref{tab:trajectorytypes} shows extended interaction in both task types. Modeling trajectories have median lengths of 54 training steps and 48 evaluation steps, compared with 40 and 38 for debugging. Median environment-tool counts are respectively 49 and 43 for modeling, and 34 and 31 for debugging. These recorded workloads support the multistep-interaction dimension described in Section~\ref{sec:sources}, complementing the business-dependency evidence in Appendix~\ref{app:cases}. They characterize execution under the reported harness rather than a policy-independent difficulty ranking.

\begin{table*}[t]

\centering\small

\setlength{\tabcolsep}{4pt}

\begin{tabular}{@{}llrrrrr@{}}
\toprule
Phase & Task type & n & Context (k tokens) & Solver steps & Env. calls & Multi-segment (\%)\\
\midrule
Training & Modeling & 1016 & 167.0 & 54.0 & 49.0 & 23.92\\
Training & Debugging & 1032 & 158.9 & 40.0 & 34.0 & 21.90\\
Evaluation & Modeling & 275 & 147.1 & 48.0 & 43.0 & 6.55\\
Evaluation & Debugging & 275 & 145.8 & 38.0 & 31.0 & 8.36\\
\bottomrule
\end{tabular}

\caption{Task-type breakdown. Context, solver steps and environment-tool calls are medians. Multi-segment denotes trajectories spanning more than one solver context.}

\label{tab:trajectorytypes}

\end{table*}

\end{document}